\documentclass[10pt,twocolumn,letterpaper]{article}

\usepackage[pagenumbers]{cvpr} 

\usepackage{colortbl} 
\usepackage{multirow}
\definecolor{cvprblue}{rgb}{0.21,0.49,0.74}
\usepackage[pagebackref,breaklinks,colorlinks,allcolors=cvprblue]{hyperref}

\def\paperID{***} 
\def\confName{CVPR}
\def\confYear{2025}

\title{Harnessing Vision Foundation Models for High-Performance, Training-Free Open Vocabulary Segmentation}

\author{
Yuheng Shi  \\
City University of Hong Kong \\
{\tt\small yuhengshi99@gmail.com}
\and
Minjing Dong  \\
City University of Hong Kong \\
{\tt\small minjdong@cityu.edu.hk}
\and
Chang Xu  \\
University of Sydney\\
{\tt\small c.xu@sydney.edu.au}
}

\begin{document}
\maketitle
\begin{abstract}
While Contrastive Language-Image Pre-training (CLIP) has advanced open-vocabulary predictions, its performance on semantic segmentation remains suboptimal. 
This shortfall primarily stems from its spatial-invariant semantic features and constrained resolution. 
While previous adaptations addressed spatial invariance semantic by modifying the self-attention in CLIP's image encoder, the issue of limited resolution remains unexplored. 
Different from previous segment-then-splice methods that segment sub-images via a sliding window and splice the results, we introduce a splice-then-segment paradigm that incorporates Segment-Anything Model (SAM) to tackle the resolution issue since SAM excels at extracting fine-grained semantic correlations from high-resolution images. 
Specifically, we introduce Trident, a training-free framework that first splices features extracted by CLIP and DINO from sub-images, then leverages SAM's encoder to create a correlation matrix for global aggregation, enabling a broadened receptive field for effective segmentation.
Besides, we propose a refinement strategy for CLIP's coarse segmentation outputs by transforming them into prompts for SAM, further enhancing the segmentation performance.
Trident achieves a significant improvement in the mIoU across eight benchmarks compared with the current SOTA, increasing from $44.4$ to $48.6$. Code is available at \url{https://github.com/YuHengsss/Trident}.

\end{abstract}    
\section{Introduction}
\label{sec:intro}
\begin{figure}[ht]
    \centering
    \includegraphics[width=0.85\linewidth]{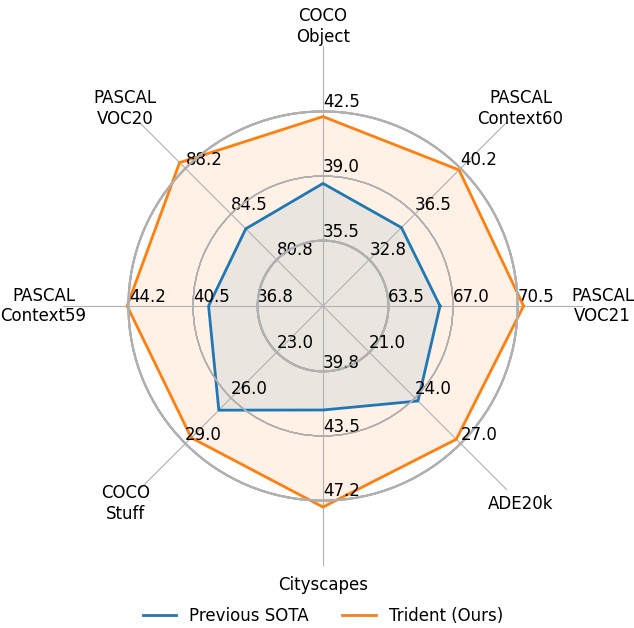}
    \vspace{-0.1in}
    \caption{Comparison with previous SOTA performance of open-vocabulary semantic segmentation under training-free setting.}
    \label{fig:overall_compare}
    \vspace{-0.2in}
\end{figure}
Semantic segmentation is a foundational vision task that aims to segment images according to different semantics, where deep learning shows impressive performance \cite{fcn,unet,badrinarayanan2017segnet,chen2017deeplab,huang2019ccnet,xie2021segformer,maskformer}.
However, these methods are trained on a close set, limiting their application in open-vocabulary scenarios. 
More recently, Vision Language Models (VLMs)~\cite{clip, cherti2023reproducible, evaclip, llava, li2022blip, li2023blipv2}, trained on web-scale data to align textual descriptions with image semantics, have demonstrated remarkable capabilities in open-vocabulary recognition. As a pioneering and representative work, CLIP~\cite{clip} has inspired numerous efforts to leverage its robust open-vocabulary capabilities for dense prediction tasks, such as semantic segmentation~\cite{maskclip,wang2024sam-clip,ovseg}.
However, CLIP only receives image-level supervision without correlating text and region-level features.
It compromises the semantic integrity of dense feature maps due to \textit{spatial invariance semantic}, indicating local features tend to be invariant to their spatial positions~\cite{wang2023sclip}, which leads to suboptimal performance of segmentation that leverages pixel-level features.
In contrast to CLIP, Vision Foundation Models (VFMs)~\cite{caron2021dino,mae,kirillov2023segment} show a close relationship between feature semantics and their positions, but they lack explicit understanding. For example, SAM segments visible objects within an image but fails to give corresponding semantics.

To migrate the spatial invariance semantic for CLIP, different adaptations are proposed and these works can be categorized into three types according to their training paradigm, \ie, full tuning~\cite{ovseg,rao2022denseclip,cho2024catseg,wang2024sam-clip}, partial tuning with additional learnable parameters~\cite{clipdenoiser,zegclip,CLIPSeg,san}, and training-free paradigm~\cite{maskclip,wang2023sclip,karazija2023ovdiff}. 
Compared to tuning-based methods, training-free paradigm offers an attractive alternative, characterized by its low cost, free of annotations, and preservation of generalization capabilities.
MaskCLIP~\cite{maskclip}, a pioneering work in this field, attributes the spatial invariance semantic of CLIP's image features to the self-attention mechanism~\cite{attention}. It proposes to replace the QK attention of the last transformer layer in CLIP's image encoder with a simple convolution, which brings significant improvements.
Inspired by this innovation, a series of studies~\cite{clipsurgery,clipdenoiser,shao2024cliptrase} have proposed more effective aggregation methods. 
However, these methods are constrained by CLIP's inherent limitation of operating at low resolutions.
To address this issue, a sliding window strategy is widely adopted to split the source image into multiple sub-images, segment them separately, and then splice the results together, which we denote as \textit{Segment-then-Splice}.
Although it alleviates this issue to some degree, it is difficult to generalize to scenarios with higher resolutions. As demonstrated in Tab.~\ref{tab: toy_size_ablation}, the performance significantly declines when the resolution increases. We mainly attribute it to the limited receptive field of the sub-images given higher resolutions. 

In this work, we propose to reformulate the sliding window strategy that improves CLIP's segmentation ability on high-resolution images. 
Motivated by the fine-grained semantic correlations of pixel-level features from SAM~\cite{kirillov2023segment}, 
we utilize a correlation matrix sourced from its encoder to harmonize the superiority of both SAM and CLIP. 
Specifically, we splice feature maps extracted from CLIP's image encoder during the sliding window procedure and incorporate the correlation matrix from SAM to further aggregate this spliced feature map for segmentation, which we denote as \textit{Splice-then-Segment}.
While feature extraction operates locally on sub-images, the correlation matrix enables global attention, extending the receptive field beyond individual windows to encompass the entire image.
To further improve the segmentation, we developed a refinement strategy. By converting CLIP's segmentation results to points, boxes, and mask prompts for SAM refinement, the results using mask prompts only can be significantly improved. 
During feature extraction of sub-images, we adopt the DINO to provide spatially covariant semantic guidance, as introduced in \cite{lan2024proxyclip}. Considering our method integrates three foundational models, \ie, CLIP, DINO, and SAM, we named it Trident.
Trident achieves SOTA performance among training-free methods, and shows competitive results even compared to weakly supervised methods~\cite{jo2024ttd,tcl} across eight widely used benchmarks, as shown in Fig.~\ref{fig:overall_compare}.

Our contributions can be summarized as: 
\textbf{(1)} We analyze the limitations of existing \textit{Segment-then-Splice} paradigm when adapting CLIP for high-resolution semantic segmentation.
\textbf{(2)} Motivated by this analysis, we introduce a \textit{Splice-then-Segment} paradigm to harmonize different vision foundation models for semantic segmentation.
\textbf{(3)} Our framework Trident is validated across 8 popular benchmarks, surpassing previous SOTA results by a significant margin.

\section{Related Works}
\label{sec:related}
\subsection{Vision Language and Foundation Models}
The introduction of Vision Language Models (VLMs), pioneered by CLIP~\cite{clip} and further developed by subsequent studies~\cite{cherti2023reproducible, evaclip, llava, li2023blipv2}, has reshaped the landscape of vision tasks. These models transitioned the field from closed-set to open-vocabulary, which is evident across various domains, including classification~\cite{guo2023calip,menon2023visual}, image captioning~\cite{yu2022coca, mokady2021clipcap} as well as detection and segmentation~\cite{ovsegmentor,ov-detr,wang2024sam-clip}. As CLIP only receives image-level supervision, the pixel-level semantic exists spatial-invariant problem~\cite{clipsurgery, maskclip} which limits its adaption to dense prediction tasks.

Another line of research in Vision Foundation Models (VFMs) focuses on learning robust vision feature representations through self-supervised learning~\cite{byol,caron2021dino,oquab2023dinov2,beit, darcet2023vision}. These VFMs exhibit strong generalization capabilities and show promising improvements in downstream tasks~\cite{li2022exploring, evaclip, chen2022vision}. For example, the DINO series~\cite{caron2021dino,oquab2023dinov2} is noted for its remarkable spatial covariant semantic representations, which enable their use in unsupervised object detection and few-shot segmentation~\cite{wang2023cut,liu2023matcher}. Recently, SAM~\cite{kirillov2023segment} introduced a VFM for image segmentation. 
The combination of innovative model design and robust data handling enabled SAM to achieve impressive zero-shot segmentation performance. 

\subsection{Open Vocabulary Semantic Segmentation}
Open vocabulary semantic segmentation aims to segment any category described through natural language. Leveraging robust open vocabulary capabilities of CLIP, many studies~\cite{rao2022denseclip,cho2024catseg,zegclip,CLIPSeg,san,maskclip,karazija2023ovdiff} have adapted its features for segmentation tasks, which could be divided into training-based and training-free approaches. Training-based methods involve introduction of additional masks~\cite{cho2024catseg,li2023open,segclip}, text annotations~\cite{cha2023learning,liu2022open,xu2022groupvit,xu2023learning} or distillation~\cite{wang2024sam-clip,wu2023clipself,clipdenoiser} which may include adjustments to CLIP’s image encoder~\cite{wang2024sam-clip,wu2023clipself} or integration of extra learnable parameters~\cite{clipdenoiser,shin2022reco,liu2024open}.
Conversely, training-free approaches avoid introducing learnable parameters and enhance CLIP's segmentation capabilities by modifying its image encoder~\cite{maskclip,wang2023sclip,clipsurgery} and utilizing VFMs~\cite{lan2024proxyclip,karazija2023ovdiff}. 

Observing the spatial invariance in CLIP's image features, MaskCLIP~\cite{maskclip} introduced a convolutional layer to replace the last Query-Key attention in CLIP's image encoder, inspiring many follow-ups~\cite{wang2023sclip,lan2024proxyclip,hajimiri2025naclip}. 
These methods typically employ a sliding window to segment each sub-image separately, addressing the limited input resolution constraint of CLIP. 
Our approach extends this by aggregating the feature map of sub-images with a global correlation matrix, which significantly improves performance.
\begin{figure*}[ht]
    \centering
    \includegraphics[width=1.0\linewidth]{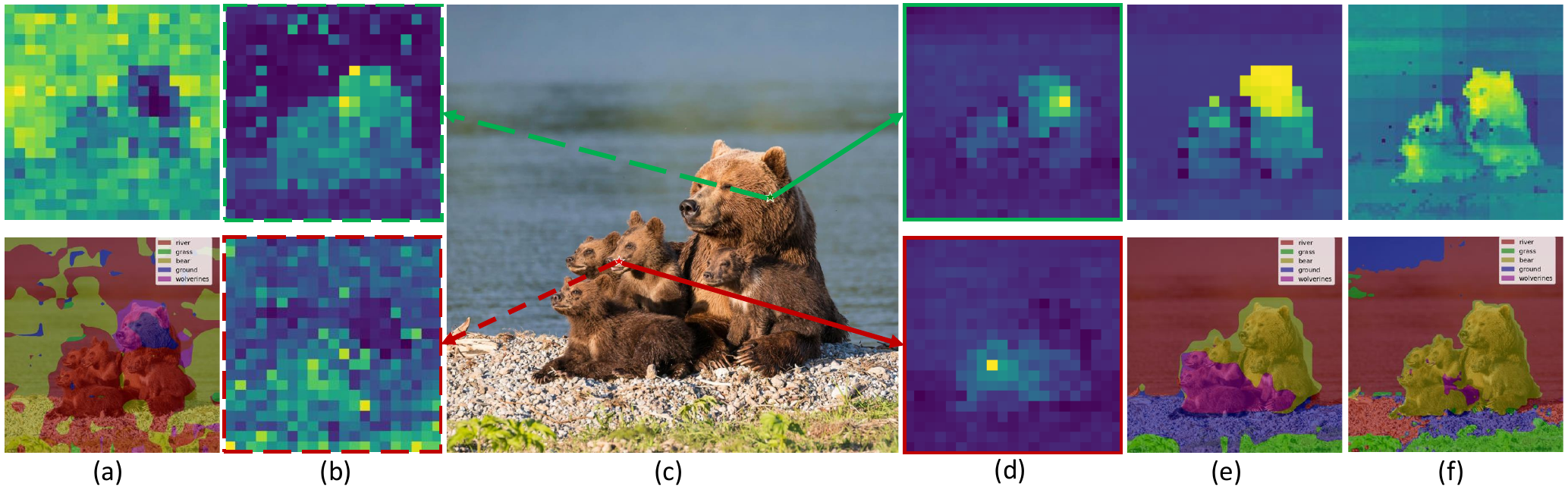}
    \vspace{-0.2in}
    \caption{Illustration of the segmentation results of CLIP and ProxyCLIP.
    Figures (a) and (e) show the results of CLIP and ProxyCLIP respectively, with an input resolution of 336 $\times$ 336. Figure (f) shows the results of ProxyCLIP with an input resolution of 1024 $\times$ 1024. The upper row of these figures shows the activation map of bear while the lower row shows the segmentation maps. Figures (b) and (d) show the attention weights and cosine similarity map in last transformer block of CLIP's image encoder and DINO's feature map respectively.
    }
    \label{fig:clip_dino_limit}
    \vspace{-0.2in}
\end{figure*}

\section{Method}
\subsection{Preliminaries}
CLIP~\cite{clip} employs a contrastive learning approach using web-scale data to simultaneously train a text encoder, $CLIP_{\text{text}}$, and an image encoder, $CLIP_{\text{img}}$. 
For a ViT-based~\cite{ViT16x16} image encoder, its feature could be denoted as  $X = \begin{bmatrix} x_{\text{cls}}, x_1, \dots, x_{HW} \end{bmatrix}\in \mathbb{R}^{(HW+1) \times d}$, where $x_{\text{cls}}$ denotes the classification token, $x_{i}$ denotes the visual token, $H$ and $W$ are the height and width of the feature map, respectively, and $d$ is the dimension of the feature space. 
To adopt CLIP for open-vocabulary segmentation, a straightforward baseline is to perform image classification on every patch. Typically, given a set of classes, $CLIP_{\text{text}}$ transforms them to text embedding $T_{\text{emd}}\in \mathbb{R}^{c \times d}$ via a fixed prompt format, where $c$ represents the number of classes.
For an image, its feature extracted from the $CLIP_{\text{img}}$ could be denoted as $I_{\text{feat}}\in \mathbb{R}^{HW \times d}$, where the $cls$ token is excluded. By calculating the cosine similarity between $T_{\text{emd}}$ and $I_{\text{feat}}$, the segmentation map $\mathcal{S}$ can be derived as:
\begin{equation}
    \label{eq: seg_procedure}
    \mathcal{S}=\underset{c}{\arg \max } \cos \left(I_{\text{feat}}, T_{\text{emd}}\right) .
\end{equation}
The skip-connection~\cite{resnet} and Feed-Forward Network (FFN) in last transformer layer of $CLIP_{\text{img}}$ are discarded following~\cite{clipsurgery,hajimiri2025naclip}.
The baseline approach (Eq.~\ref{eq: seg_procedure}) yields suboptimal results due to the spatially invariant nature of attention mechanisms and dense features in CLIP's image encoder. 
We provide visualization for the segmentation results and the attention weights of given points in Fig.~\ref{fig:clip_dino_limit} (a) and (b) respectively, to facilitate intuitive understanding.

\subsection{Analysis for Segment-then-Splice Paradigm}
\label{sec:analysis_previous_paradigm}
Although previous works~\cite{maskclip,wang2023sclip,hajimiri2025naclip} have improved the localization ability of CLIP for segmentation, they still utilize \textit{Segment-then-Splice} paradigm to migrate CLIP for higher resolution images:
\begin{equation}
\scalebox{0.91}{$ \displaystyle
\label{eq:img_split and res_splice}
    \begin{aligned}
        & \relax [I_{\text{src}}^1, I_{\text{src}}^2, \cdots, I_{\text{src}}^n] = \sigma(I_{\text{src}}),\; I_{\text{feat}}^i = \text{LP}(A^iV^i),\\
        & \mathcal{S}^{i}=\underset{c}{\arg \max } \cos \left(I^{i}_{\text{feat}}, T_{\text{emd}}\right), \;
        \mathcal{S} = \gamma([\mathcal{S}^1, \mathcal{S}^2, \cdots, \mathcal{S}^n]),\\
    \end{aligned}
$}
\end{equation}
where $\sigma$ represents the sliding window transformation that divides the source image $I_{\text{src}}$ into $n$ sub-images, $\gamma$ denotes the splicing operation for segmentation fragment $\mathcal{S}^i$ by Eq.~\ref{eq: seg_procedure} and $\text{LP}$ denotes linear projection. 
The feature extraction for sub-image $I_{\text{src}}^i$ through $CLIP_{\text{img}}$ remains similar across methods up to the penultimate layer. We only show the key distinction in how they process the attention value $V^i$ in the final layer through their respective correlation terms $A^i$, \ie, MaskCLIP~\cite{maskclip} employs a near-identity convolution, and ProxyCLIP~\cite{lan2024proxyclip} utilizes DINO-based cosine similarity.
In the following analysis, we take ProxyCLIP as the example due to SOTA performance.

\begin{table}
\centering
\caption{Performance along input resolution of ProxyCLIP with DINO-B/16 on Pascal VOC dataset. The size denotes the shorter side resolution of the source images. Resolution for sliding window is 224 $\times$ 224 when the input size is 224 and 336 for others.}
\label{tab: toy_size_ablation}
\setlength{\tabcolsep}{0.05cm} 
\begin{tabular}{@{}l|ccccc|ccccc@{}}
\toprule
\textbf{Setting}& \multicolumn{5}{c}{\textbf{Without Background}} & \multicolumn{5}{c}{\textbf{With Background}} \\
\midrule
\textbf{Size} & 224 & 336 & 448 & 576 & 688 & 224 & 336 & 448 & 576 & 688 \\ 
\midrule
\textbf{mIoU} & 79.3 & 79.7 & 78.5 & 73.4 & 70.0 & 56.3 & 59.3 & 59.2 & 56.1 & 53.5 \\
\bottomrule
\end{tabular}
\vspace{-5mm}
\end{table}
Our experiments on the PASCAL VOC dataset~\cite{voc12} reveal a non-monotonic relationship between input resolution and segmentation performance, as documented in Tab.~\ref{tab: toy_size_ablation}. This phenomenon, consistently observed across multiple settings, can be explained by analyzing the interplay between source image resolution and receptive field coverage of sub-images.
When the source image resolution $R_{src}$ is comparable to the sliding window resolution $R_{sub}$, as in the cases of 224 and 336 pixels, each sub-image's receptive field encompasses nearly the entire image. Under these conditions, the paradigm effectively leverages increased resolution for fine-grained segmentation. 
However, as $R_{src}$ increases while maintaining fixed $R_{sub}$ (without considering stride overlap), the number of sub-images $n=R_{src}/R_{sub}$ grows linearly while the relative receptive field of each sub-image inversely decreases.
When a sub-image's receptive field becomes insufficient to encompass entire objects, CLIP's classification capability is significantly impaired. 
Specifically, increasing shorter side resolution of source image from 336 to 688 pixels results in mIoU decreases of 9.7\% and 5.8\% for settings with and without background, respectively.
Qualitative analysis in Fig.~\ref{fig:clip_dino_limit} (f) further supports these findings. While higher resolution provides fine-grained results compared to the 336-pixel setting in Fig.~\ref{fig:clip_dino_limit} (e), the limited receptive field manifests in reduced classification at bear centers and river boundaries. The activation maps of bear clearly reveal these windowing artifacts, demonstrating the fundamental limitations of the Segment-then-Splice approach at high resolutions.

Given the quadratic complexity of attention and the lack of training on high-resolution images, further increasing the input resolution for CLIP does not significantly enhance performance and results in a substantial increase in computational costs~\cite{wu2023clipself}. 
These observations underscore the need for further aggregation of these sub-image's feature maps, particularly for source images at higher resolutions.

\subsection{Splice-then-Segment Paradigm}
\label{sec: sam correlation}
\begin{figure}[t]
    \centering
    \vspace{-3mm}
    \includegraphics[width=1.0\linewidth]{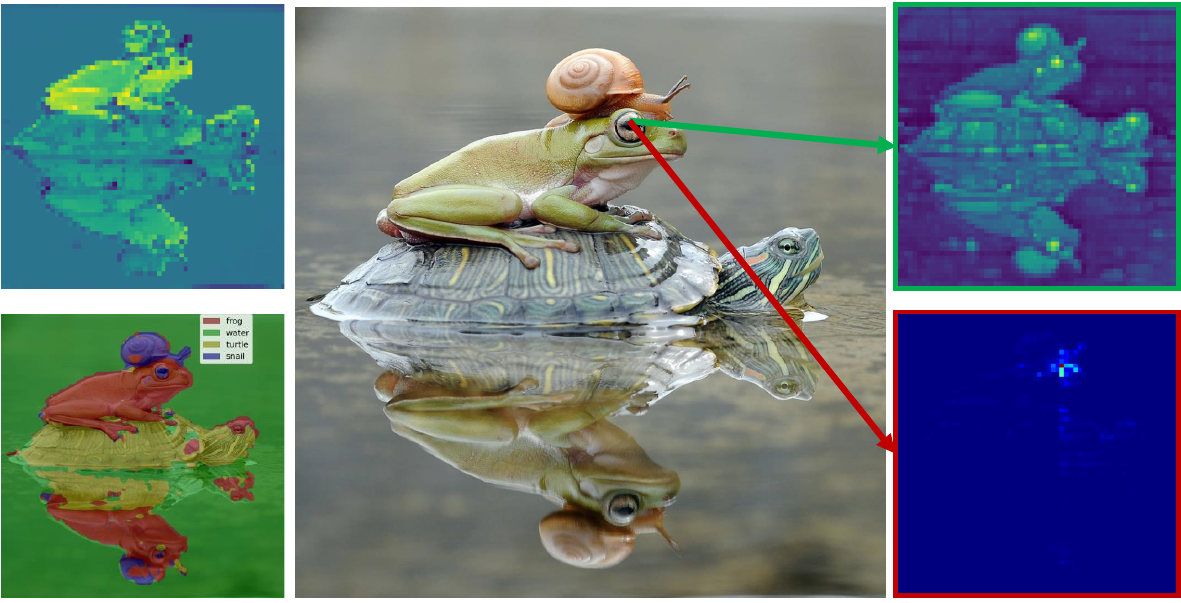}
    \caption{Segmentation results using our Splice-then-Segment paradigm. 
     Left: activation map (top) and segmentation results (bottom) for the frog class. Right: cosine similarity map (top) and attention map (bottom) for the given point.}
    \label{fig:sam_limit}
    \vspace{-3mm}
\end{figure}

\begin{figure*}[htb]
    \centering
    \includegraphics[width=1.0\linewidth]{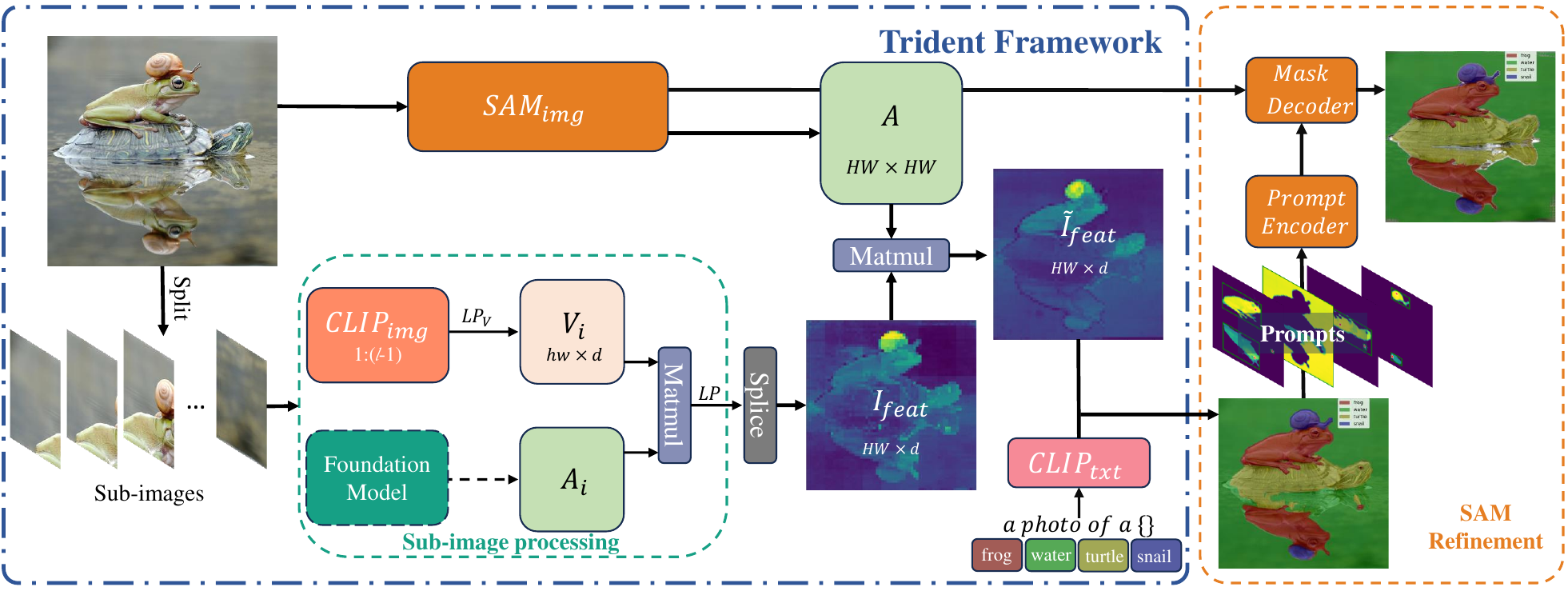}
    \caption{Framework of the proposed Trident model. Foundation models are first used to introduce correlations for sub-image's features. Subsequently, a correlation matrix derived from the source image and SAM is utilized to aggregate features across different sub-images. The resulting segmentation maps can then serve as prompts for further refinement by SAM.}
    \label{fig:framework}
    \vspace{-5mm}
\end{figure*}

\noindent \textbf{Framework Overview.} To overcome the limitations inherent in processing sub-images independently, we introduce a \textit{Splice-then-Segment} paradigm by reformulating Eq.~\ref{eq:img_split and res_splice}:
\begin{equation}
\begin{aligned}
    \label{eq:feature_splice}
    I_{\text{feat}} &= \gamma([I_{\text{feat}}^1, I_{\text{feat}}^2, \cdots, I_{\text{feat}}^n]), \quad
    \tilde{I}_{\text{feat}} = AI_{\text{feat}}, \\
    \mathcal{S} &= \underset{c}{\arg \max} \, \cos(\tilde{I}_{\text{feat}}, T_{\text{emd}}).
\end{aligned}
\end{equation}
Here, feature maps of sub-images are first spliced together by $\gamma$ defined in Eq.~\ref{eq:img_split and res_splice} to form an integral feature map $I_{\text{feat}}$. 
Subsequently, this feature map undergoes global aggregation through correlation matrix $A$, producing $\tilde{I}_{\text{feat}}$ with enhanced cross-window contextual information.
For brevity, the interpolation for $I_{\text{feat}}$ to align its size with $A$ is omitted here. Finally, $\tilde{I}_{\text{feat}}$ is utilized to generate segmentation results as described in Eq.~\ref{eq: seg_procedure}. 
The key distinction from Eq.~\ref{eq:img_split and res_splice} lies in our global correlation matrix approach. While previous methods apply correlation matrices independently to each sub-image's features, our method enables semantic aggregation across all sub-images simultaneously. This global aggregation effectively extends the receptive field from individual sub-images to the entire source image, thereby addressing CLIP's inherent resolution limitations.

For effective feature aggregation, the correlation matrix $A$ should fulfill dual objectives: capturing semantic relationships between features in $I_{\text{feat}}$ to enable intra-class aggregation, while maintaining fine-grained correlations through high-resolution feature processing. Given these requirements, we leverage SAM~\cite{kirillov2023segment} for correlation matrix generation, as it uniquely processes high-resolution image, addressing resolution limitations common to VFMs.


\noindent \textbf{Correlation Matrix.}  The correlation matrix can be initially constructed using the cosine similarity of SAM's features $F$, with an additional masking mechanism $M$ adopted from ProxyCLIP~\cite{lan2024proxyclip}. This formulation can be expressed as:
\begin{equation}
\scalebox{0.94}{$ \displaystyle
\begin{aligned}
\label{eq: naive correlation}
     & F = SAM_{\text{img}}(I_{\text{src}}), \quad
     C = \frac{F}{\|F\|} \left(\frac{F}{\|F\|}\right)^T, \quad  \\
     & A = \text{Softmax}(C + M),
     \text{where } M_{ij} = \begin{cases}
                            0, &  C_{ij} \geq \epsilon, \\
                            -\infty, &  C_{ij} < \epsilon.
                            \end{cases}
\end{aligned}
$}
\end{equation}
The parameter $\epsilon$ serves as a threshold. In Fig.~\ref{fig:sam_limit}, we present the results using our Splice-then-Segment paradigm and the correlation matrix defined in Eq.~\ref{eq: naive correlation}. Specifically, the top figure in the first column displays the activation map for the frog class. 
Our Splice-then-Segment paradigm eliminates the window panel effect of previous methods while achieving superior segmentation coherence.

While leveraging SAM's feature-based cosine similarity for the correlation matrix helps mitigate the limited receptive field issue, the correlation coefficients $A_{i,j}$ in Eq.~\ref{eq: naive correlation} may not optimally capture semantic relationships at the predefined category level.
As SAM is designed to segment any visible object, it may over-segment predefined objects, a phenomenon also observed in SAM-CLIP~\cite{wang2024sam-clip}. 
This limitation is evident in the cosine similarity matrix $C$, where individual points primarily capture low-level visual features rather than object-level semantic relationships characteristic of DINO (Fig.~\ref{fig:clip_dino_limit} (d)). As illustrated in Fig.~\ref{fig:sam_limit}, the similarity map for a point on a frog's eye exhibits high correlation with visually similar structures (e.g., snail centers and turtle eyes) due to shared local patterns, but fails to establish strong correlations with the complete frog object, thereby compromising classification accuracy.

To alleviate the limitations of correlation matrix derived from SAM's feature, a higher level of semantic correlation is required. 
An alternative involves utilizing the attention weights $W$ from SAM's last encoder layer. These attention weights demonstrate enhanced semantic correlation, as shown in Fig.~\ref{fig:sam_limit}, but they also inevitably incorporate background features, as both foreground and background features are necessary for effective segmentation. 
To address this limitation, we propose an improved aggregation matrix $A$ that combines cosine similarity correlation $C$ with attention weights $W$:
\begin{equation}
    \label{eq: refined_guidance}
    A = \frac{W + M}{\|W + M\|}, M_{ij} = \begin{cases}
                            0, &  C_{ij} \geq \epsilon, \\
                            -W_{ij}, & C_{ij} < \epsilon.
              \end{cases}
\end{equation}
Here, $\epsilon$ and $C$ are defined similarly to those in Eq.~\ref{eq: naive correlation}. 
This formulation selectively preserves attention weights only for token pairs whose cosine similarity exceeds the threshold $\epsilon$, effectively suppressing attention to likely background elements. We term this selective correlation matrix the affinity matrix. Empirical results demonstrate that this hybrid approach consistently outperforms methods using either cosine similarity or attention weights in isolation.

\noindent \textbf{Trident.} The limitations of SAM's low-level semantic correlations during global aggregation can be further mitigated by incorporating more robust, high-level features during sub-image processing.
Therefore, a better correlation term when extracting sub-image's feature, as stated in Eq.~\ref{eq:img_split and res_splice}, is also helpful. Consequently, we also preserve DINO~\cite{caron2021dino} to provide object-level spatially covariant semantic correlation during the feature extraction of sub-images.
Our proposed Trident framework, illustrated in Fig.~\ref{fig:framework}, synergistically combines three components: CLIP for fundamental semantic representation, DINO for object-level correlation in sub-images, and SAM for global feature aggregation.

\subsection{SAM Refinement}
\label{sec:sam_refine}
To achieve more fine-grained results and fully utilize SAM, segmentation outputs from Trident can serve as prompts for further refinement by SAM's prompt encoder and mask decoder. 
However, using segmented masks only as prompts for the prompt encoder leads to unsatisfactory results, as SAM is primarily trained with points and boxes as prompts~\cite{kirillov2023segment}. To address this, we transform the segmentation results into point, box, and mask prompts using a strategy described below. Employing these three types of prompts simultaneously enhances the refinement quality.
For $k_{th}$ class, we generate a binary segmentation map $\mathcal{B}_{k}$ from the segmentation result $\mathcal{S}$:
\begin{equation}
    \mathcal{B}_{k}(x,y) = \begin{cases}
        1, & \text{if } \mathcal{S}(x,y) = k, \\
        0, & \text{otherwise},
    \end{cases}
\end{equation}
where $(x,y)$ denotes pixel coordinates. Subsequently, $\mathcal{B}_{k}$ is decomposed into $n_k$ connected regions $\{\mathcal{R}^i_k\}_{i=1}^{n_k}$ via morphology methods~\cite{wu2005optimizing,fiorio1996two}. For each region, point prompt $p^i_k$, box prompt $b^i_k$, and mask prompt $m^i_k$ are defined as:
\begin{equation}
\scalebox{1.0}{$ \displaystyle
\begin{aligned}
    & p^i_k = \underset{(x,y) \in \mathcal{R}^i_k}{\arg \max} \, \mathcal{M}_k(x,y), \\
    & b^i_k = \text{bbox}(\mathcal{R}^i_k), \quad m^i_k = \alpha \mathcal{B}_k \cdot \mathcal{M}_k,
\end{aligned}
$}
\end{equation}
where $\mathcal{M}_k$ represents the confidence scores from Trident for class $k$, $\text{bbox}(\cdot)$ denotes the minimal axis-aligned bounding box operator that returns coordinates enclosing the input region, and $\alpha$ is a scaling coefficient.
Empirical results suggest setting $\alpha$ to a small value (e.g., 0.005) to optimize the effectiveness of the mask prompt, as confirmed by our ablation studies.
Due to SAM's architectural constraint of processing single-class mask prompts, the refinement of segmentation masks is performed independently for each semantic category. After obtaining the segmented score from SAM, we multiply it by $\mathcal{M}$ to generate the final results. The SAM refinement process and prompt examples is depicted in the right part of Fig.~\ref{fig:framework} for clear visualization (best viewed in zoom-in).

\begin{table*}[ht]
\setlength{\tabcolsep}{0.15cm}
\centering
\caption{\textbf{Quantitative Comparison of Recent Open Vocabulary Segmentation Works.} The highest-performing result is highlighted in \textbf{bold}, and the second highest in \underline{underline} for clarity. Results marked with a $\dagger$ are cited from ProxyCLIP~\cite{lan2024proxyclip}.}
\label{tab:segmentation_comparison}
\vspace{-2mm}
\begin{tabular}{@{}lcccccccccc@{}}
\toprule
\multirow{3}{*}{\textbf{Method}}& \multirow{3}{*}{\textbf{Train}} & \multicolumn{3}{c}{\textbf{With background}} & \multicolumn{5}{c}{\textbf{Without background}} & \multirow{3}{*}{\textbf{Avg.}} \\ 
\cmidrule(lr){3-5} \cmidrule(lr){6-10}
& & \textbf{VOC21} & \textbf{Context60} & \textbf{Object} & \textbf{VOC20} & \textbf{Context59} & \textbf{Stuff} & \textbf{City} & \textbf{ADE} \\
\midrule
SAM-CLIP\textcolor{gray}{\textsubscript{[CVPR'24]}} & \checkmark & 60.6 & 29.2 & - & - & - & \textbf{31.5} & - & 17.1 & - \\ 
CoDe\textcolor{gray}{\textsubscript{[CVPR'24]}} & \checkmark & 57.7 & 30.5 & 32.3 & - & - & 23.9 & 28.9 & 17.7& - \\
TTD\textcolor{gray}{\textsubscript{[ECCV'24]}} & \checkmark & 61.1 & 37.4 & 37.4 & - & - & 23.7 & 27.0 & 17.0 & - \\
TCL\textcolor{gray}{\textsubscript{[CVPR'23]}} & \checkmark & 51.2 & 24.3 & 30.4 & 77.5 & 30.3 & 19.6 & 23.5 & 14.9 & 33.9 \\
CLIP-DINOiser\textcolor{gray}{\textsubscript{[ECCV'24]}} & \checkmark & 62.1 & 32.4 & 34.8 & 80.9 & 35.9 & 24.6 & 31.7 & 20.0 & 40.3  \\
\midrule
\rowcolor{Black!10!white} \multicolumn{11}{l}{\textcolor{Black}{\textbf{CLIP ViT-B/16}}} \\
CLIP\textcolor{gray}{\textsubscript{[ICML'21]}}     &  & 16.4 & 8.4  & 5.6  & 41.9 & 9.2 & 4.4 & 5.0 & 2.9 & 11.7 \\
MaskCLIP\textcolor{gray}{\textsubscript{[ECCV'22]}} &  & 43.4 & 23.2  & 20.6  & 74.9 & 26.4 & 16.7 & 24.9 & 11.9 & 30.3 \\
ReCo\textcolor{gray}{\textsubscript{[NeurIPS'22]}}&  & 25.1 & 19.9& 15.7& 57.7& 22.3& 14.8& 21.6 & 11.2& 23.5 \\
OVDiff\textcolor{gray}{\textsubscript{[ECCV'24]}}&  & 66.3 & 29.7& 34.6& 80.9& 32.9& 20.3& 23.4 & 14.1& 37.8 \\
SCLIP\textcolor{gray}{\textsubscript{[ECCV'24]}} &  & 59.1 & 30.4& 30.5& 80.4& 34.2& 22.4& 32.2& 16.1& 38.2 \\
NACLIP\textcolor{gray}{\textsubscript{[WACV'25]}}&  & 58.9 & 32.2& 33.2& 79.7& 35.2& 23.3& 35.5& 17.4& 39.4 \\
ProxyCLIP\textcolor{gray}{\textsubscript{[ECCV'24]}}&  & 61.3 & 35.3& 37.5& 80.3& 39.1& 26.5& 38.1 & 20.2& 42.3 \\
\rowcolor{SkyBlue!20!white}
Trident (Ours) &  & \underline{67.1} &  \underline{38.6}&  \underline{41.1}&  \underline{84.5}&  \underline{42.2}&  28.3 &  \underline{42.9} & 21.9&  \underline{45.8} \\
\midrule
\rowcolor{Black!10!white} \multicolumn{11}{l}{\textcolor{Black}{\textbf{OpenCLIP ViT-H/14}}} \\
OpenCLIP\textcolor{gray}{\textsubscript{[CVPR'23]}}     &  & 8.8 & 5.0  & 5.3  & 21.7 & 5.5 & 3.2 & 4.3 & 2.8 & 7.1 \\
MaskCLIP$^{\dagger}$\textcolor{gray}{\textsubscript{[ECCV'22]}} &  & 31.4 & 13.3  & 16.2  & 41.7 & 15.8 & 8.4 & 17.7 & 10.4 & 19.3 \\
SCLIP$^{\dagger}$\textcolor{gray}{\textsubscript{[ECCV'24]}} &  & 43.8 & 23.5  & 24.6  & 67.5 & 25.6 & 16.8 & 19.5 & 11.3 & 29.1 \\
ProxyCLIP\textcolor{gray}{\textsubscript{[ECCV'24]}}&  & 65.0 & 35.4& 38.6& 83.3& 39.6& 26.8& 42.0 & \underline{24.2}& 44.4 \\
\rowcolor{SkyBlue!20!white}
Trident (Ours) &  & \textbf{70.8} & \textbf{40.1}& \textbf{42.2}& \textbf{88.7}& \textbf{44.3}& \underline{28.6}& \textbf{47.6} & \textbf{26.7}& \textbf{48.6} \\
\bottomrule
\end{tabular}
\end{table*}

\section{Experiments}

\subsection{Benchmark Settings.}
\label{sec: bench_settings}
Following established practices~\cite{maskclip, wang2023sclip, lan2024proxyclip, tcl}, we evaluate our approach using six widely-used segmentation datasets: PASCAL VOC 2012 (VOC)~\cite{voc12}, PASCAL Context (Context)~\cite{pascalcontext}, COCO Object (Object)~\cite{lin2014microsoftcoco}, COCO Stuff (Stuff)~\cite{coco}, Cityscapes (City)~\cite{cityscapes}, and ADE20k (ADE)~\cite{ade20k}. The abbreviations in parentheses denote the short names of the datasets.
Specifically, the PASCAL VOC 2012 and PASCAL Context datasets are subdivided into two settings—VOC20 and VOC21 for the former, and Context59 and Context60 for the latter, based on the inclusion of background elements. 
Benefiting from the splice-then-segment paradigm, Trident can utilize higher resolutions to enhance performance. To accommodate different dataset configurations, we resize images accordingly: the shorter side is set to 336 pixels for VOC20, 448 pixels for VOC21, Object, and Stuff, 576 pixels for Context59, Context60, and ADE, and 688 pixels for Cityscapes. 
All benchmarks use a sliding window of 336 $\times$ 336 pixels and the stride is 224 for most, except for VOC20, which uses a stride of 112. For the textual prompt construction, we utilize a standard ImageNet manner following previous works~\cite{clip, cherti2023reproducible, clipdenoiser}. As the proposed method is training-free, we report the mean Intersection over Union (mIoU) on the validation split of mentioned benchmarks directly. The implementation is based on MMSegmentation~\cite{mmseg2020} framework.

\noindent \textbf{Baselines and Comparison Methods.} We establish our baselines using CLIP~\cite{clip} with ViT-B/16 and OpenCLIP~\cite{cherti2023reproducible} with ViT-H/14 as the simplest models. Additionally, MaskCLIP~\cite{maskclip} and ProxyCLIP~\cite{lan2024proxyclip} serve as advanced baselines. 
Beyond these, we consider recent SOTA open vocabulary semantic segmentation methods, including training-free approaches like ReCo~\cite{shin2022reco}, OVDiff~\cite{karazija2023ovdiff}, SCLIP~\cite{maskclip}, CLIPtrase~\cite{kang2024lavg}, NACLIP~\cite{hajimiri2025naclip}, and training-based methods such as SAM-CLIP~\cite{wang2024sam-clip}, TCL~\cite{tcl}, CoDe~\cite{CoDe}, TTD~\cite{jo2024ttd}, and CLIP-DINOiser~\cite{clipdenoiser}. All results reported here do not involve pixel-level post-processing methods like PAMR~\cite{araslanov2020pamr}.
We follow the implementation details from ProxyCLIP to evaluate the performance of the vanilla CLIP baseline. For other competitors, we report their performance as described in their respective publications, unless otherwise specified.

\subsection{Main Results}

\begin{figure*}[htb]
    \centering
    \includegraphics[width=0.98\linewidth]{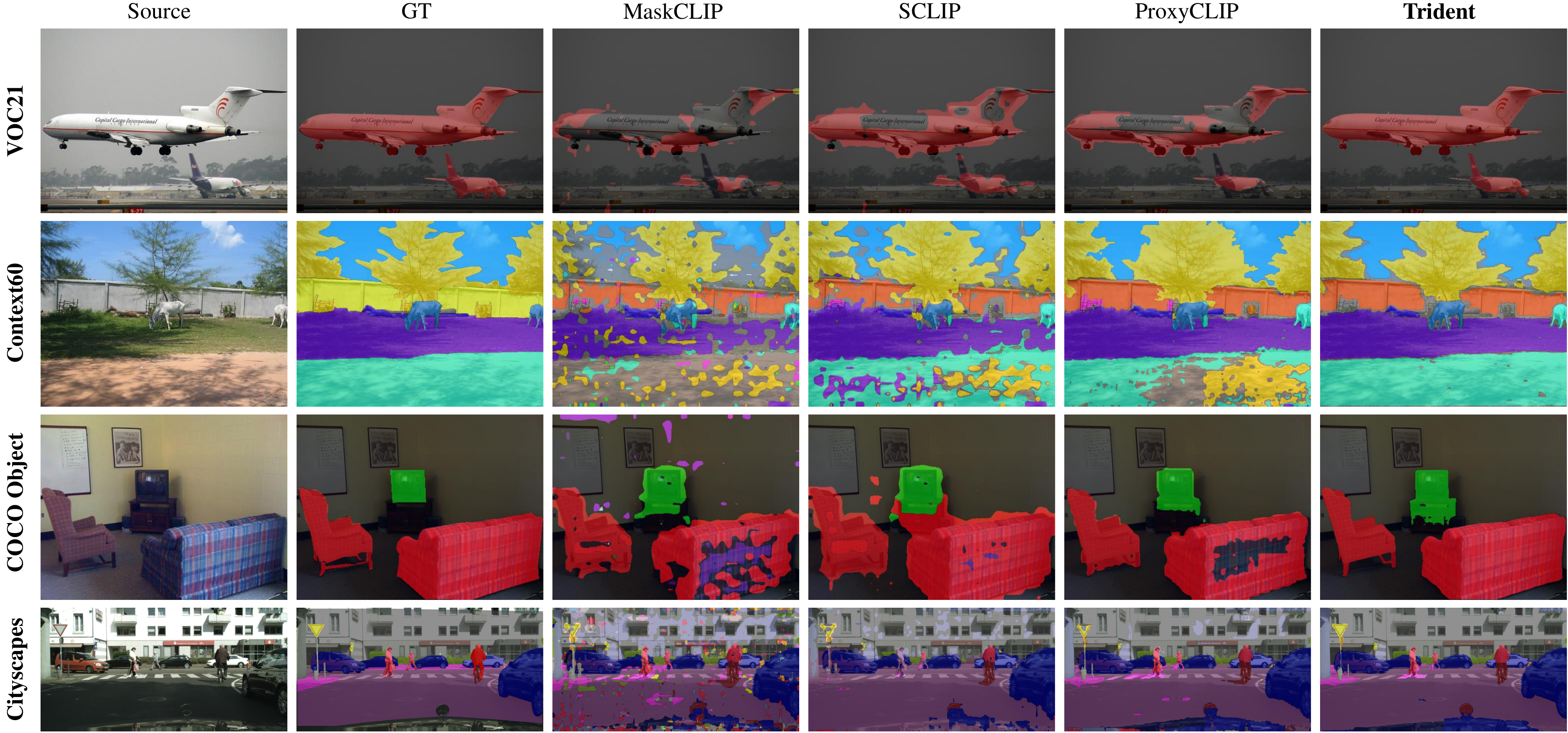}
    \vspace{-2mm}
    \caption{\textbf{Qualitative comparison with previous training-free open vocabulary segmentation methods.}}
    \label{fig:vis compare}
    \vspace{-5mm}
\end{figure*}

\noindent \textbf{Quantitative Comparison}. 
Tab.~\ref{tab:segmentation_comparison} presents comprehensive performance comparisons between our Trident framework and existing approaches. Among methods utilizing CLIP ViT-Base architecture, Trident consistently outperforms both training-based and training-free approaches across seven benchmarks. Notably, in the training-free category, Trident achieves SOTA performance across all eight benchmarks, with significant margins over previous best results: \textbf{+3\%} mIoU on PASCAL Context~\cite{pascalcontext} and COCO Object~\cite{lin2014microsoftcoco}, and \textbf{+4\%} mIoU on CityScapes~\cite{cityscapes}. Compared to the previous SOTA method, ProxyCLIP~\cite{lan2024proxyclip}, Trident demonstrates a substantial average improvement of \textbf{3.5\%} mIoU across all benchmarks.
Trident maintains its superior performance when compared to training-based methods, demonstrating an average improvement of \textbf{5\%} mIoU over CLIP-DINOiser~\cite{clipdenoiser}. While SAM-CLIP~\cite{wang2024sam-clip} achieves higher performance on COCO Stuff~\cite{coco} (31.5\% vs. our 28.3\%), this advantage comes at the cost of extensive training on multiple large-scale datasets~\cite{cc3m,cc12M,yfcc15m,ridnik2021imagenet}, which compromises its open-vocabulary capabilities. This trade-off is evident in SAM-CLIP's performance degradation on other benchmarks, where it lags behind our method by approximately 6\% mIoU on PASCAL VOC and ADE20K.
The advantages of Trident become more pronounced when implemented with OpenCLIP ViT-Huge architecture. Our framework surpasses previous SOTA results by substantial margins: \textbf{+5\%} mIoU on both PASCAL VOC~\cite{voc12} and CityScapes~\cite{cityscapes} benchmarks, with an average improvement of \textbf{4.2\%} mIoU across all benchmarks.

\noindent \textbf{Qualitative Comparison}. 
Fig.~\ref{fig:vis compare} presents qualitative comparisons between Trident and previous SOTA methods, all utilizing the CLIP-B/16 architecture. The visual results highlight Trident's superior performance in two aspects: improved semantic consistency in object recognition and more precise segmentation boundary delineation. These advantages are particularly pronounced in complex scenes from Context60 and Cityscapes benchmarks.

\begin{table}[!ht]
\small
\setlength{\tabcolsep}{0.10cm}
\centering
\caption{\textbf{Ablation for Different Paradigm and Correlation Matrix on MaskCLIP.} Corr. denotes the type of correlation matrix.}
\label{tab:paradigm-ablation}
\begin{tabular}{@{}lccccccccc@{}}
\toprule

\textbf{Corr.} & \textbf{V21} & \textbf{C60} & \textbf{Obj.} & \textbf{V20} & \textbf{C59} & \textbf{Stf.} & \textbf{City} & \textbf{ADE} &  \textbf{Avg.} \\
\midrule
\rowcolor{Black!10!white} \multicolumn{10}{l}{\textcolor{Black}{\textbf{Segment-then-Splice}}} \\
None & 47.2 & 24.5 & 26.4 & 63.6 & 26.9 & 18.8 & 28.2 & 13.3 & 31.1 \\
Cos & 48.1 & 29.6 & 29.1 & 73.2 & 32.6 & 21.6 & 22.7 & 15.8 & 34.1 \\
\midrule
\rowcolor{Black!10!white} \multicolumn{10}{l}{\textcolor{Black}{\textbf{Splice-then-Segment}}} \\
Cos & 59.3 & 34.8 & 35.2 & 77.5 & 37.9 & 25.0 & 36.1 & 18.8 & 40.6 \\
Attn & 61.7 & 35.7 & 38.4 & 83.1 & 39.2 & 26.0 & 37.2 & 20.1 & 42.7 \\
Aff & 62.8 & 36.5 & 38.6 & 82.8 & 39.9 & 26.5 & 37.3 & 20.3 & 43.1 \\
\bottomrule
\end{tabular}
\end{table}

\begin{table}
\vspace{-2mm}
\small
\setlength{\tabcolsep}{0.07cm}
\centering
\caption{\textbf{Ablation for Component of Trident upon ProxyCLIP baseline.} Aff. and Ref. denote affinity matrix and SAM refinement, respectively.}
\label{tab:trident-ablation}
\begin{tabular}{@{}lcccccccccc@{}}
\toprule

\textbf{Aff.} &  \textbf{Ref.} & \textbf{V21} & \textbf{C60} & \textbf{Obj.} & \textbf{V20} & \textbf{C59} & \textbf{Stf.} & \textbf{City} & \textbf{ADE} &  \textbf{Avg.} \\
\midrule
\rowcolor{Black!10!white} \multicolumn{11}{l}{\textcolor{Black}{\textbf{CLIP-B/16 + SAM-B/16}}} \\
            &            & 59.2 &33.8 &  35.4&	79.7  & 37.0  & 25.7& 38.5& 19.0 & 41.0 \\
 \checkmark &            & 64.5 &37.2 &	39.5&	83.7 &	40.9 & 27.6& 40.4& 20.9 & 44.3 \\
 \checkmark & \checkmark        & 67.1 &38.6 &	41.1&	84.5 &	42.2 & 28.3& 42.9& 21.9 & 45.8 \\
\checkmark & \textcolor{gray}{PAMR} & \textcolor{gray}{63.4} & \textcolor{gray}{36.8} & \textcolor{gray}{38.6} & \textcolor{gray}{82.9} & \textcolor{gray}{40.6} & \textcolor{gray}{27.7} & \textcolor{gray}{40.7} & \textcolor{gray}{20.8} & \textcolor{gray}{43.9} \\
\midrule
\rowcolor{Black!10!white} \multicolumn{11}{l}{\textcolor{Black}{\textbf{ OpenCLIP-H/14 + SAM-H/16}}} \\
            &            & 62.2 &34.3 &	 37.3&	 83.7 &	 38.0 & 26.4& 40.9& 23.6 & 43.3 \\
 \checkmark &            & 69.3 &38.5 &	41.0&	87.8 &	42.9 & 27.7& 44.2& 25.6 & 47.1 \\
 \checkmark & \checkmark & 70.8 &40.1 &	42.2&	88.7 &	44.4 & 28.6& 47.6& 26.7 & 48.6 \\
\bottomrule
\end{tabular}
\vspace{-5mm}
\end{table}

\subsection{Ablation Study}
\label{sec: ablation}
In all ablation studies, we use ViT-B/16 as the default setting for foundation models including CLIP, DINO, and SAM, unless otherwise noted. The input resolution for different benchmarks inherent from Sec.~\ref{sec: bench_settings}. Thus, performance may exists differences with main table for baseline methods. The VOC21, Context60, COCO Object, VOC20, Context59, and Stuff benchmarks may be abbreviated as V21, C60, Obj., V20, C59, and Stf., respectively, to conserve space in some ablation tables.

\noindent \textbf{On the Splice-then-Segment Paradigm}. To evaluate the effectiveness of the proposed Splice-then-Segment paradigm, we begin with the MaskCLIP~\cite{maskclip} baseline, which does not introduce any further aggregation for the sub-image's feature maps, to exclude the impact of other factors. Note SAM refinement is not included here.
The detailed ablation results for the effects of different paradigms and various correlation matrices $A$ are presented in Tab~\ref{tab:paradigm-ablation}. 
We evaluate four variants of correlation matrices: (1) None, corresponding to the MaskCLIP baseline without feature correlation; (2) Cos, utilizing cosine similarities of SAM's features as defined in Eq.~\ref{eq: naive correlation}; (3) Attn, employing attention weights from SAM's final transformer block; and (4) Aff, our proposed affinity matrix detailed in Eq.~\ref{eq: refined_guidance}. Note that in the Segment-then-Splice paradigm, Cos matrices are computed and applied independently for each sub-image's feature map.
When utilizing cosine similarity to construct the correlation matrix, our proposed Splice-then-Segment outperforms the previous paradigm by over 6\% mIoU on average. Switching from cosine similarity to attention weights yields an additional 2.1\% mIoU improvement, supporting our analysis of SAM feature limitations. The integration of our proposed affinity matrix further enhances performance from 42.7\% to 43.1\% mIoU. Overall, compared to the MaskCLIP baseline, our Splice-then-Segment paradigm with affinity matrix achieves a substantial improvement of 12.0\% mIoU on average.

\noindent \textbf{On the Component of Trident}. 
We conduct comprehensive ablation studies on ProxyCLIP with DINO-B/16 as a stronger baseline. Starting from the Segment-then-Splice paradigm, we progressively incorporate our Splice-then-Segment paradigm with affinity matrix and SAM refinement. The results are detailed in Tab.~\ref{tab:trident-ablation}.
Our framework's scalability is evaluated across different model configurations. For CLIP ViT-B/16, we pair it with SAM-Base to balance computational efficiency and performance. For OpenCLIP-H/14, we employ SAM-Huge to explore the framework's performance ceiling. The Splice-then-Segment paradigm with affinity matrix yields improvements of 3.3\% and 3.8\% mIoU for base and huge versions, respectively, with SAM refinement contributing an additional 1.5\% mIoU improvement in both cases.
We also compare SAM refinement with the widely adopted PAMR post-processing~\cite{araslanov2020pamr} in the base configuration. Unlike previous CLIP-based methods that produce noisy segmentation maps, our approach generates significantly cleaner results. Consequently, PAMR post-processing not only fails to provide additional benefits but leads to a 0.4\% mIoU performance degradation.

\begin{table}
\setlength{\tabcolsep}{0.1cm}
\centering
\caption{\textbf{Ablation for Prompts in SAM Refinement.}}
\label{tab:sam-refine-ablation}
\begin{tabular}{@{}lcccccccc@{}}
\toprule
\textbf{Mask} & \textbf{Box} & \textbf{Point} & \textbf{V21}   & \textbf{C60} & \textbf{Obj.}  & \textbf{Stf.} & \textbf{City.} & \textbf{ADE.}  \\
\midrule
          &            &           &	64.5  &37.2     &  39.5  &27.6 &40.4 &20.9\\
\checkmark&            &           &  58.3  &33.8     &33.0	   &25.9 &34.4 &19.9\\
& \checkmark & \checkmark          &  58.9  &34.3     &	36.4   &25.5 &34.8 &19.2 \\
\checkmark& \checkmark &\checkmark &  67.1  &38.6     &	41.1   &28.3 &42.9 &21.9\\
\bottomrule
\end{tabular}
\vspace{-2mm}
\end{table}

\begin{table}
\setlength{\tabcolsep}{0.18cm}
\centering
\caption{\textbf{Ablation for $\alpha$ Coefficient in Mask Prompt.} The default setting is highlighted in \textbf{bold}.}
\label{tab:sam-coff-ablation}
\begin{tabular}{@{}lccccccc@{}}
\toprule

\textbf{$\alpha$} & 1.0 & 0.5 & 0.1   & 0.01 & \textbf{0.005}  & 0.003 & 0.001  \\
\midrule
\textbf{V20}    & 78.7   & 79.9&	83.3  &84.2 &  84.5  &84.5 &84.4\\
\textbf{City.}  & 25.8   & 29.4&	36.8  &41.4 &  42.9  &42.9 &42.4\\
\bottomrule
\end{tabular}
\end{table}

\begin{table}
\setlength{\tabcolsep}{0.06cm}
\centering
\caption{\textbf{Ablation for Different Input Resolution.} SAM refinement is excluded here in Trident to clearly show the effect of the proposed framework. The setting is an abbreviation of \textbf{shorter side} - \textbf{window size} - \textbf{stride}. }
\label{tab:size-ablation}
\begin{tabular}{@{}lccccc@{}}
\toprule

\textbf{Setting} & \textbf{VOC20} & \textbf{Context59} & \textbf{Context60} & \textbf{Stuff} & \textbf{ADE}  \\
\midrule
\rowcolor{Black!10!white} \multicolumn{6}{l}{\textcolor{Black}{\textbf{ProxyCLIP}}} \\
336-336-112& 79.7 &34.4 &	38.1&	25.7 &	19.4  \\
448-336-224& 78.5 &34.8 &	38.3&	25.7 &	19.3 \\
576-336-224& 73.4 &33.8 &	37.0&	24.6 &	19.0 \\

\midrule
\rowcolor{Black!10!white} \multicolumn{6}{l}{\textcolor{Black}{\textbf{Trident w.o SAM Refine}}} \\
336-336-112& 83.7 &35.8 &	39.6&	27.0 &	20.7  \\
448-336-224& 83.3 &37.0 &	40.7&	27.6 &	20.6  \\
576-336-224& 82.0 &37.2 &	40.9&	27.5 &	20.9 \\
\bottomrule
\end{tabular}
\vspace{-5mm}
\end{table}

\noindent \textbf{On SAM Refinement}.
Tab.~\ref{tab:sam-refine-ablation} presents our ablation study on SAM refinement prompt types. Using individual prompt types yields suboptimal results, performing worse than the unrefined baseline. However, the synergistic combination of all three prompt types consistently improves performance across all benchmarks, achieving an average gain of 1.6\% over unrefined results. We further analyze the impact of the mask prompt scaling coefficient $\alpha$ on VOC20 and Cityscapes benchmarks as shown in Tab.~\ref{tab:sam-coff-ablation}. As discussed in Sec.~\ref{sec:sam_refine}, setting $\alpha=1$ (no scaling) significantly degrades performance below the unrefined baseline. Performance improves substantially as $\alpha$ decreases logarithmically, reaching optimal results at $\alpha=0.005$, which we adopt as our default configuration for SAM refinement.

\noindent \textbf{On the Effect of Input Resolution}. To verify the claim that Segment-then-Splice paradigm might diminish with increased resolution, we conducted ablation studies on more datasets for ProxyCLIP, and our Trident. The detailed results are presented in Table~\ref{tab:size-ablation}. ProxyCLIP exhibits deteriorating performance as resolution increases; specifically, its performance on the VOC20 dataset decreases from 79.7\% to 73.4\% mIoU when the input resolution is increased from 336 to 576 for the shorter side of the image. This trend is consistent across other datasets, as the receptive field becomes more constrained in the Segment-then-Splice paradigm with increased resolution. In contrast, our Splice-then-Segment paradigm benefits from increased resolution, thereby obtaining better performance with increased resolution. A notable exception is the VOC20 dataset, where the performance of our method also declines due to baseline’s significantly reduced effectiveness.

\begin{table}
\setlength{\tabcolsep}{0.12cm}
\centering
\caption{\textbf{Efficiency Analysis for Trident Framework.} The inference latency is tested on RTX 4090 GPU with FP16 precision.}
\vspace{-3mm}
\label{tab:eff-ablation}
\begin{tabular}{@{}lccccc@{}}
\toprule

\textbf{DINO} & \textbf{SAM} & \textbf{Ref.} & \textbf{\#Params.}   & \textbf{Mem.} & \textbf{Thru.}  \\
             &               &               &\textbf{(M)}          &   (MB)        & (imgs/sec) \\
\midrule
\rowcolor{Black!10!white} \multicolumn{6}{l}{\textcolor{Black}{\textbf{CLIP-B/16 + SAM-B/16}}} \\
          &            &           &	149  &\phantom{1}672     &  118.8  \\
\checkmark&            &           &  234  &	\phantom{1}851 &	\phantom{1}68.5 \\
\checkmark& \checkmark &           &  323  &	2501           &	\phantom{1}15.3 \\
\checkmark& \checkmark &\checkmark &  364 &	2526               &	\phantom{1}10.0 \\

\midrule
\rowcolor{Black!10!white} \multicolumn{6}{l}{\textcolor{Black}{\textbf{OpenCLIP-H/14 + SAM-H/16}}} \\
          &            &           &	\phantom{1}986  &2308     &  28.2  \\
\checkmark&            &           &  1071            &2484     &	22.4 \\
\checkmark& \checkmark &           &  1708            &	5804    &	\phantom{1}6.4 \\
\checkmark& \checkmark &\checkmark &  1749            &	5827    &	\phantom{1}5.0 \\
\bottomrule
\end{tabular}
\vspace{-5mm}
\end{table}

\subsection{Efficiency Analysis.}
As the proposed Trident integrates three foundational models, we conducted an analysis of their impact on inference costs. We adopted an image resolution of 448 $\times$ 448 and utilized a sliding window with a size of 336 and stride of 224 for CLIP and DINO. For SAM, the input resolution was set to 1024. The throughput was tested on an RTX 4090 GPU using FP16 precision for all models. The detailed results are reported in Tab.~\ref{tab:eff-ablation}, which includes results for both the base and huge versions of these models. The introduction of SAM resulted in a significant increase in GPU memory usage and processing time, primarily due to its demand for high-resolution inputs. Additionally, the incorporation of SAM refinement led to a slight increase in both GPU memory usage and time cost.

\section{Conclusion}
We present Trident, a training-free framework that advances open-vocabulary semantic segmentation through the strategic integration of CLIP, DINO, and SAM. The core innovation lies in our Splice-then-Segment paradigm, which replaces the Segment-then-Splice approach with a more effective global feature aggregation strategy. By leveraging SAM-derived correlation matrices, our method successfully addresses CLIP's inherent limitations in processing high-resolution images. The framework is complemented by a refinement mechanism that enhances segmentation accuracy. Comprehensive evaluations across eight benchmarks demonstrate its superior performance over existing training-free methods, marking a significant advancement in open-vocabulary semantic segmentation.

{
    \small
    \bibliographystyle{ieeenat_fullname}
    \bibliography{main}
}


\end{document}